\documentclass[]{spie}  %

\usepackage{amsmath,amsfonts,amssymb}
\usepackage{graphicx}
\usepackage[colorlinks=true, allcolors=blue]{hyperref}

\usepackage{cleveref}
\usepackage{siunitx}
\usepackage{booktabs}
\usepackage{tabularx}

\title{On Background Bias in Deep Metric Learning}

\author{Konstantin Kobs}
\author{Andreas Hotho}
\affil{University of Würzburg, Am Hubland, 97074 Würzburg, Germany}

\authorinfo{Further author information: (Send correspondence to K.K.)\\K.K.: E-mail: kobs@informatik.uni-wuerzburg.de\\  A.H.: E-mail: hotho@informatik.uni-wuerzburg.de}

\begin{document} 
\maketitle

\begin{abstract}
Deep Metric Learning trains a neural network to map input images to a lower-dimensional embedding space such that similar images are closer together than dissimilar images.
When used for item retrieval, a query image is embedded using the trained model and the closest items from a database storing their respective embeddings are returned as the most similar items for the query.
Especially in product retrieval, where a user searches for a certain product by taking a photo of it, the image background is usually not important and thus should not influence the embedding process.
Ideally, the retrieval process always returns fitting items for the photographed object, regardless of the environment the photo was taken in.
In this paper, we analyze the influence of the image background on Deep Metric Learning models by utilizing five common loss functions and three common datasets.
We find that Deep Metric Learning networks are prone to so-called background bias, which can lead to a severe decrease in retrieval performance when changing the image background during inference.
We also show that replacing the background of images during training with random background images alleviates this issue.
Since we use an automatic background removal method to do this background replacement, no additional manual labeling work and model changes are required while inference time stays the same.
Qualitative and quantitative analyses, for which we introduce a new evaluation metric, confirm that models trained with replaced backgrounds attend more to the main object in the image, benefitting item retrieval systems.
\end{abstract}
 

\section{INTRODUCTION}
\label{sec:intro}

\begin{figure}[t]
    \centering
    \includegraphics[width=\textwidth]{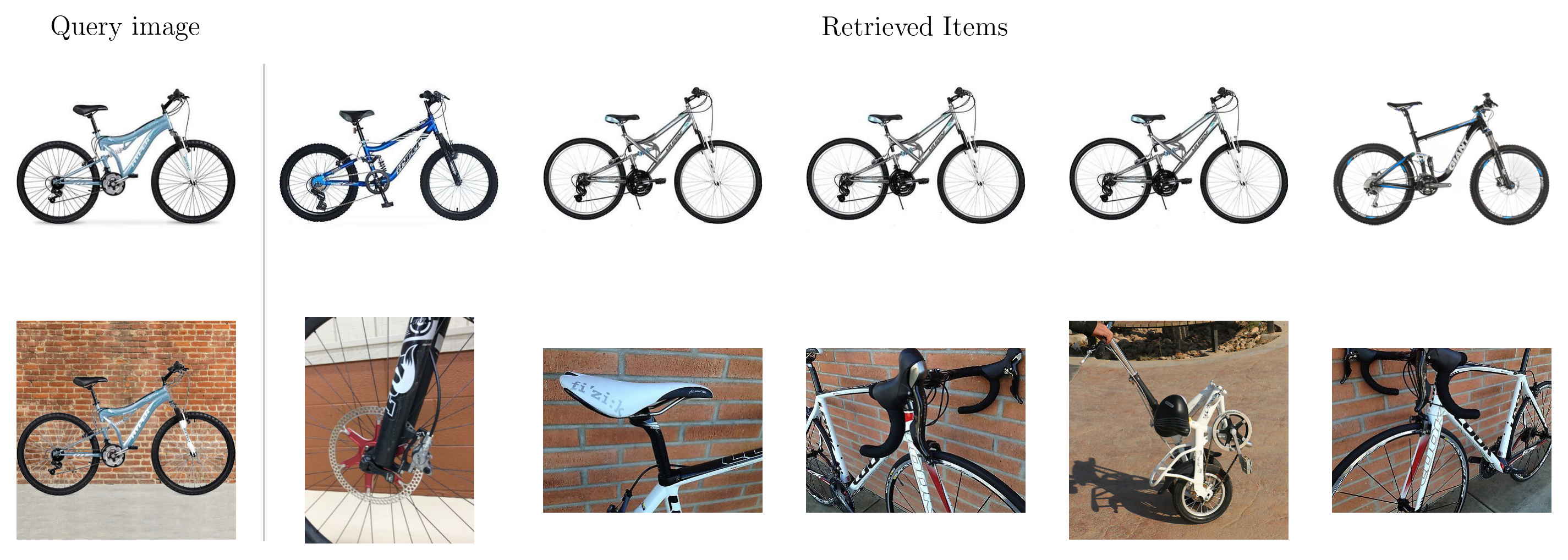}
    \caption{Retrieval results for two query images (first column) based on the distance of embeddings of a DML model trained on the Stanford Online Products~\cite{songDeepMetricLearning2016} dataset with the Contrastive loss~\cite{hadsellDimensionalityReductionLearning2006}. The second query image shows the exact same object as the first one, but we exchange the background using an image editing software. Ideally, the embeddings for the two images should be similar since they show the same object, leading to similar retrieval results. However, both queries result in very different retrieval results mostly based on background similarity. While the first row only shows images that have a white background, the second one only shows images with patterns resembling the brick wall background in the query image. This behavior is not desirable in item retrieval systems. In this paper, we investigate the influence of the background on the retrieval performance of DML models.}
    \label{fig:figure_1}
\end{figure}

Deep Metric Learning (DML) is the task of learning a neural network to embed input items (in this case, images) such that embeddings of similar items are closer together than embeddings of dissimilar items~\cite{musgraveMetricLearningReality2020}.
This technique is often used for face recognition, person reidentification, and item retrieval~\cite{kaya2019deep}.
For instance in item retrieval, a query image of an item is used to find semantically similar images by identifying the closest images in embedding space.
Two images are deemed similar if they show the same item.
Given this definition, the background of the images should not play a role in the embedding process, since objects can be photographed in different environments and thus appear in front of different backgrounds.
Similar desired properties can be defined for other DML applications such as person reidentification.

Previous analytical work for the different task of \textbf{content classification} shows that neural networks suffer from so-called \textit{background bias}, i.e. they use information from the image background to identify the image category.
For example, image classifiers trained to identify ships often focus on the water and not on the ship itself.
This way, the classifier is not able to identify ships at land~\cite{lapuschkin2019unmasking}.

Since DML does not classify images but embeds them, the findings on background bias from the literature are not directly transferable to these models.
If background bias was also present in DML models, image backgrounds would influence the embedding process.
Then, taking a picture of an object on the street or in a studio setup could lead to different search results when searched for in item retrieval methods, resulting in performance degradations of the item retrieval system.
\Cref{fig:figure_1} shows such a situation:
Placing the bike in front of a brick wall or a studio backdrop gives completely different nearest neighbor search results.
This is not desirable, since the retrieval system should only take the main object into account.

In this paper, we investigate background bias in DML by conducting multiple experiments on three standard DML datasets (Cars196~\cite{krause3DObjectRepresentations2013}, CUB200~\cite{wahCaltechUCSDBirds2002011Dataset}, Stanford Online Products~\cite{songDeepMetricLearning2016}) and five different DML loss functions.
We design a test setting where we replace image backgrounds with other images and measure the retrieval's performance drop compared to the unmodified images; larger drops in performance indicate that the model relies more on the background.
We show that, depending on the dataset, models can suffer from severe background bias.
To combat this behavior, we apply a simple but effective training strategy that does require no additional manual labeling work or model changes and keeps the same inference times.
For this, we extract the main object from the images during training using a \textit{salient object detection method}~\cite{Qin_2020_PR} and put them onto randomly selected background images.
We show that this technique, which we call BGAugment, indeed improves performance in our test setting, even though no foreground/background segmentation is available during testing, indicating that the model learns to focus less on the background.
To verify this, we qualitatively and quantitatively analyze the resulting models and show that the model trained with BGAugment attends more to the main object instead of the background, leading to better performance when backgrounds change.
For this, we introduce a metric that quantifies the focus of the model on the foreground and background.\footnote{Our code is available at \url{https://github.com/LSX-UniWue/background-bias-in-dml}}

Our contributions in this paper are threefold:

\begin{itemize}
    \item We are the first to show that DML models suffer from background bias, depending on the dataset
    \item We apply a simple but effective method to alleviate background bias in DML for item retrieval that does not require additional labeling work, model changes, or increases in inference time
    \item We compare and analyze models trained using both methods qualitatively and quantitatively using input attribution methods and propose a new metric that quantifies the focus of a model on the foreground
\end{itemize}

\section{RELATED WORK}

In recent years, a large corpus of literature has investigated background bias in \textbf{classification} neural networks.
They find that neural networks often use indicators from the background of images, such as the environment, to identify the correct class for a given input image.
While during test time, only new images from a fixed set of classes are given for classification, in a typical DML setting, the test classes are disjoint from the training classes~\cite{musgraveMetricLearningReality2020}.
Also, DML networks map images to an $n$-dimensional embedding space and do not classify them.
Thus, findings of background bias in classification models do not directly transfer to the DML setting.

In addition, methods developed to combat background bias are specialized to classification networks and cannot be directly applied to DML networks.
Such methods can be divided into two categories, which we term Background Augmentation and Attribution Regularization.
Background Augmentation methods exchange the background of images during training or inference with random images~\cite{xiao2020noise,tian2018eliminating,kc2021impacts}.
This way, the model cannot find correlations between background features and class labels.
Another work proposes to crop the image near the main object to prevent background being visible in the image~\cite{wen2022fighting}.
In our experiments, we use a Background Augmentation technique.
Attribution Regularization computes the attribution map of an input sample during training to identify the image regions the model focuses on.
The loss function then guides the model to produce attribution maps that resemble the image's foreground/background segmentation map~\cite{ross2017right,shao2021right,liu2019exploring,chefer2022optimizing}.
While attribution map generation methods for DML models exist~\cite{kobs2021different}, Attribution Regularization has not been applied yet to DML.

Related fields of background bias are also investigated.
Neural network classifiers often suffer from simplicity bias~\cite{shah2020pitfalls}, using the simplest clues to classify an image.
Training an additional network that complements a biased model~\cite{nam2020learning} or ensembles that learn diverse feature sets alleviate the problem that the model only learns a few potentially irrelevant features~\cite{pagliardini2022agree}.
To prevent models from using spurious correlations between the image and the class label~\cite{yang2022understanding,szyc2021checking,sagawa2020investigation}, the network's last layer can be fine-tuned on data that does not show such correlations~\cite{kirichenko2022last}.

Kobs et al.~\cite{kobs2021different} investigate the influence of different image factors such as item or background color on different DML models.
For this, they generate fake car images using 3D rendering software in a controlled way and measure the change in performance for DML models trained on Cars196~\cite{krause3DObjectRepresentations2013}.
While this approach can investigate the influence of different factors, it is limited to image datasets which can be generated in a controlled way, which is often tedious work.
Our method is bound to the investigation of background bias in DML, but can be applied to all image datasets.
We additionally apply a simple but effective strategy to alleviate background bias.

\section{METHODOLOGY}

In this section, we introduce our new test setting that quantifies the dependence of trained DML models on the image background.
Intuitively, the more a DML model attends to the background of images to generate an embedding, the larger the change of the embeddings when changing the image's background.
In turn, when randomly changing the background of test images in the DML setting, the retrieval performance should drop substantially if the model pays much attention to the background.

\subsection{Test Setting}

We can assume that in item retrieval, the most salient object in an image is the object that was intended to be photographed.
Thus, for each image $I \in \mathbb{R}^{h \times w \times 3}$ (RGB image with width $w$ and height $h$) in the test dataset $\mathcal{D}$, we identify the main object and create a binary mask $M \in [0,1]^{h \times w}$ separating the most salient object from the background ($1$/white denotes main object, $0$/black denotes background).
In order to obtain such masks, we use the salient object detection neural network U$^2$-Net~\cite{Qin_2020_PR}.
It is designed to detect the most salient regions --- in our case the main object --- in the image and outputs a binary mask that separates the object from the background.
We verify the segmentation quality of the network by computing the average overlap of generated and hand-annotated masks on a randomly sampled subset of images for each tested dataset.
Overlap is defined as the percentage of the ground truth foreground area that is also covered by the generated mask.
We use overlap as a metric since it is more important to cover the relevant parts of the image than removing the background.
On average, the automatic mask generator has an overlap of more than \SI{90}{\%} with the manual annotated binary masks for all datasets, so the generated masks mostly cover the relevant parts of the image.

In addition to the image $I$ and mask $M$, we collect a dataset of background images scraped from the popular stock photo website Unsplash\footnote{Scraped from \url{https://unsplash.com/s/photos/background} on May 13th, 2022}.
We filter the dataset such that no obvious foreground objects are present in the images.
The resulting background image dataset $\mathcal{B}$ contains one hundred images and shows solid colors, color gradients, background objects, and abstract patterns.

During testing, for each image $I$, we sample a background photo $B \in \mathcal{B}$ to create a new test example $I'$ with
\begin{equation}
    I'_{:,:,c} = M \odot I_{:,:,c} + (\mathbf{1} - M) \odot I_{:,:,c}
\end{equation}
for each color channel $c \in \{\text{red}, \text{green}, \text{blue}\}$.
Here, $\odot$ is the element-wise multiplication and $\mathbf{1} = 1^{h \times w}$, i.e. a matrix with the same size as $M$ consisting of ones.
We call the newly created test dataset $\mathcal{D}'$ the ``corrupted'' test set, while the original test dataset $\mathcal{D}$ is called the ``clean'' test set.
Both datasets $\mathcal{D}$ and $\mathcal{D}'$ are fed separately through the trained model and a common evaluation metric is computed for the resulting embeddings.
Here we use Mean Average Precision at R (MAP@R)~\cite{musgraveMetricLearningReality2020}.
We can then observe the drop in performance from the clean to the corrupted dataset.
In order to average out the influence of background samples, we run our test setting five times and report means and standard deviations.

\subsection{BGAugment: Background Replacement During Training}

In order to combat background bias in Deep Metric Learning, we apply a simple but effective strategy during training inspired by the literature about background bias in classification networks.
We call this method BGAugment.
Similar to the test setting description, we replace the background of input images.
However, we do this during training and validation.
To not interfere with the background images used in the test set, we collect another one hundred background images from Unsplash.

During each training iteration, we sample a random background image for each training image and use the automatically generated binary mask from the salient object detection model to replace the image background.
Since we use a pretrained salient object detection model~\cite{Qin_2020_PR}, there is no need for additional manual data labeling.
Also, since we do not apply the background replacement during inference, there is \textit{no computational overhead when applying the model in production}.
Additionally, BGAugment only touches the data loading process, leaving all other training components such as the model or loss function intact, allowing for fast adoption of this technique.

\section{EXPERIMENTAL SETUP}

\subsection{Loss Functions}

In our experiments, we compare five different loss functions.
Kobs et al. have shown that there are differences in the features learned by networks trained using these loss functions, depending on whether they are ranking or classification based~\cite{kobs2021different}.
We thus evaluate overall five loss functions: the three ranking losses Contrastive Loss~\cite{hadsellDimensionalityReductionLearning2006}, Triplet Loss~\cite{weinbergerDistanceMetricLearning2006}, and Multi Similarity Loss~\cite{wangMultiSimilarityLossGeneral2019}, as well as the two classification losses ArcFace Loss~\cite{dengArcFaceAdditiveAngular2019} and Normalized Softmax Loss~\cite{liuSphereFaceDeepHypersphere2018,wangNormFaceL2Hypersphere2017,zhaiClassificationStrongBaseline2019}.
This should allow us to identify differences in their performances.

\subsection{Datasets}

We perform experiments on three standard benchmark datasets for Deep Metric Learning: Cars196, CUB200, and Stanford Online Products.
\textbf{Cars196}~\cite{krause3DObjectRepresentations2013} consists of car images of overall \num{196} car models.
The first half of classes is used to train a DML model, the second \num{98} car models are used for testing.
The cars are mostly photographed in different locations, lighting conditions, and from different angles.
\textbf{CUB200}~\cite{wahCaltechUCSDBirds2002011Dataset} shows \num{200} different bird species.
Again, the dataset is split into two halves for training and testing.
The images show the birds in their natural environments.
\textbf{Stanford Online Products (SOP)}~\cite{songDeepMetricLearning2016} contains images of products from \textit{ebay}.
Each class in SOP consists of all images from one product page, thus often showing the same product from different angles.

\subsection{Training and Evaluation Setup}

We mostly follow the training procedure described by Musgrave et al. who design a fair setting to compare different DML loss functions~\cite{musgraveMetricLearningReality2020}.
As a model base, we use a BNInception~\cite{ioffeBatchNormalizationAccelerating2015a} network pretrained on ImageNet~\cite{imagenet} with frozen Batch Normalization layers.
The last fully connected layer is replaced to output \num{128} dimensional vectors.
The outputs are normalized to unit length to stabilize training.
The model is trained with a learning rate of \num{e-6} on the first \SI{80}{\%} of training classes and validated on the remaining \SI{20}{\%}.
Musgrave et al. report the best hyperparameters for each loss function on each dataset by conducting a cross validation.
We adopt these hyperparameters for our experiments and use them for all tested methods, i.e. there is no difference in hyperparameters for the BGAugment runs.
While a dedicated hyperparameter search might improve the performance, we want to investigate how well BGAugment performs when just applied to an existing model setup.
More information on the training process can be found in the original paper by Musgrave et al~\cite{musgraveMetricLearningReality2020}.

\section{RESULTS}

\begin{table}[t]
    \centering
    \caption{Mean and standard deviations of MAP@R for our experiments. All values are given in percent. The best results between the vanilla and BGAugment models are written in bold.}
    
    \begin{tabular}{lcc|cc|cc}
        \toprule
        {} & \multicolumn{2}{c}{\textbf{Cars196}} & \multicolumn{2}{c}{\textbf{CUB200}} & \multicolumn{2}{c}{\textbf{SOP}} \\
        {} & \textbf{clean} & \textbf{corrupted} & \textbf{clean} & \textbf{corrupted} & \textbf{clean} & \textbf{corrupted} \\
        \midrule
        \textbf{Contrastive}        & 15.22 & 13.31 ± 0.15 & \textbf{20.27} & 12.47 ± 0.08 & \textbf{37.98} & 4.17 ± 0.05 \\
        \textbf{+ BGAugment}        & \textbf{16.43} & \textbf{16.31 ± 0.03} & 19.39 & \textbf{15.46 ± 0.08} & 31.52 & \textbf{24.09 ± 0.04} \\
        \midrule
        \textbf{Triplet}            & \textbf{15.33} & 13.40 ± 0.04 & 18.29 & 10.94 ± 0.05 & \textbf{36.71} & 6.05 ± 0.06 \\
        \textbf{+ BGAugment}        & 15.28 & \textbf{15.19 ± 0.05} & \textbf{18.56} & \textbf{15.33 ± 0.16} & 29.63 & \textbf{21.39 ± 0.07} \\
        \midrule
        \textbf{Multi Similarity}   & \textbf{18.80} & \textbf{16.25 ± 0.06} & \textbf{19.19} & 12.21 ± 0.16 & \textbf{39.65} & 6.51 ± 0.06 \\
        \textbf{+ BGAugment}        & 15.61 & 15.48 ± 0.05 & 18.93 & \textbf{15.65 ± 0.16} & 32.67 & \textbf{24.00 ± 0.07} \\
        \midrule
        \textbf{ArcFace}            & \textbf{16.25} & 13.63 ± 0.08 & 20.66 & 12.87 ± 0.16 & \textbf{40.50} & 6.98 ± 0.05 \\
        \textbf{+ BGAugment}        & 16.19 & \textbf{16.23 ± 0.04} & \textbf{20.91} & \textbf{17.92 ± 0.08} & 22.25 & \textbf{16.53 ± 0.05} \\
        \midrule
        \textbf{Normalized Softmax} & 18.06 & 15.36 ± 0.03 & 20.18 & 12.41 ± 0.10 & \textbf{41.52} & 7.67 ± 0.04 \\
        \textbf{+ BGAugment}        & \textbf{18.15} & \textbf{18.12 ± 0.05} & \textbf{20.60} & \textbf{17.39 ± 0.10} & 32.16 & \textbf{24.60 ± 0.03} \\
        \bottomrule
    \end{tabular}
    \label{tab:results}
\end{table}

\Cref{tab:results} shows the means and standard deviations of the Mean Average Precision at R (MAP@R)~\cite{musgraveMetricLearningReality2020} for our experiments.
Depending on the dataset, the drop in performance from the clean to the corrupted test set can be severe.
While for Cars196, the performance drops by only around \num{2} to \num{3} percentage points for all loss functions, CUB200 and SOP show much larger differences (approx. \num{8} and \num{34} percentage points, respectively).
That gives a relative performance drop of around \SI{20}{\%}, \SI{40}{\%}, and \SI{85}{\%} for Cars196, CUB200, and SOP, respectively.
Even though the performance of all models on the corrupted dataset is still better than randomly sampling embeddings, the drop in performance is substantial.
We hypothesize that this is due to the training datasets' properties.
The images in Cars196 show cars in different environments, so the background does not often correlate with the similarity between images.
Since the birds in the CUB200 dataset are shown in their natural habitat, the environment gives clues about the similarity between images.
For example, there are waterbirds and landbirds in the dataset, thus the background features can be used to differentiate between them.
The background influence is the most severe for the ebay product images in the SOP dataset.
Images of one product are often taken in the same environmental conditions.
The DML model then picks up these features to embed the image, since they are similar across images of one class and thus can be used to find similarities between the images.

Between loss functions, we observe no large difference in drops and overall performance on all datasets, indicating similar vulnerability to background bias of ranking and classification based losses.
The use of BGAugment improves the models' performance on the corrupted dataset, except for the Multi Similarity Loss on the Cars196 dataset.
On Cars196 and CUB200, BGAugmented models perform similarly or even outperform their base model on the clean dataset.
This means that the backgrounds of images in Cars196 and CUB200 are not necessary to achieve good performance.

On the other hand, applying BGAugment to models trained on SOP improves the performance on the corrupted dataset significantly but shows large performance drops on the clean dataset.
We hypothesize that the high performance without BGAugment is only achievable by exploiting the background.
In other words, the good performance of models on the clean SOP dataset is misleading in terms of item retrieval, since the models are not able to keep up the performance when backgrounds are exchanged during training.
In realistic item retrieval settings, query images most often show other backgrounds than the images in the database.

\section{ANALYSIS}

\begin{figure}[t]
    \centering
    \includegraphics[width=0.48\textwidth]{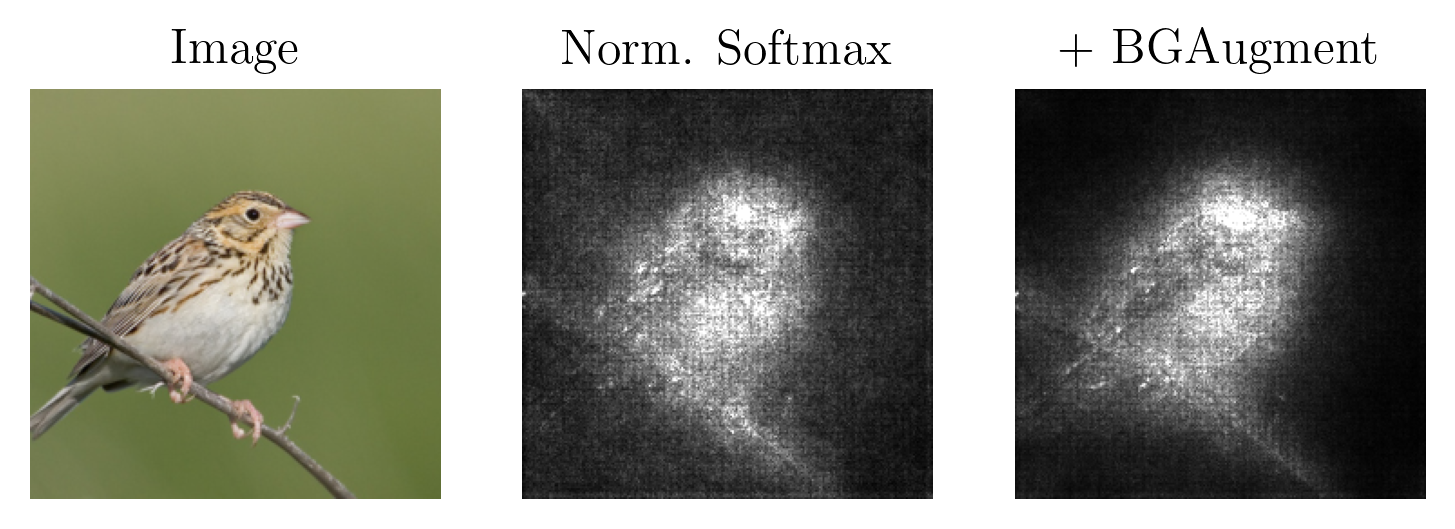}
    \hfill
    \includegraphics[width=0.48\textwidth]{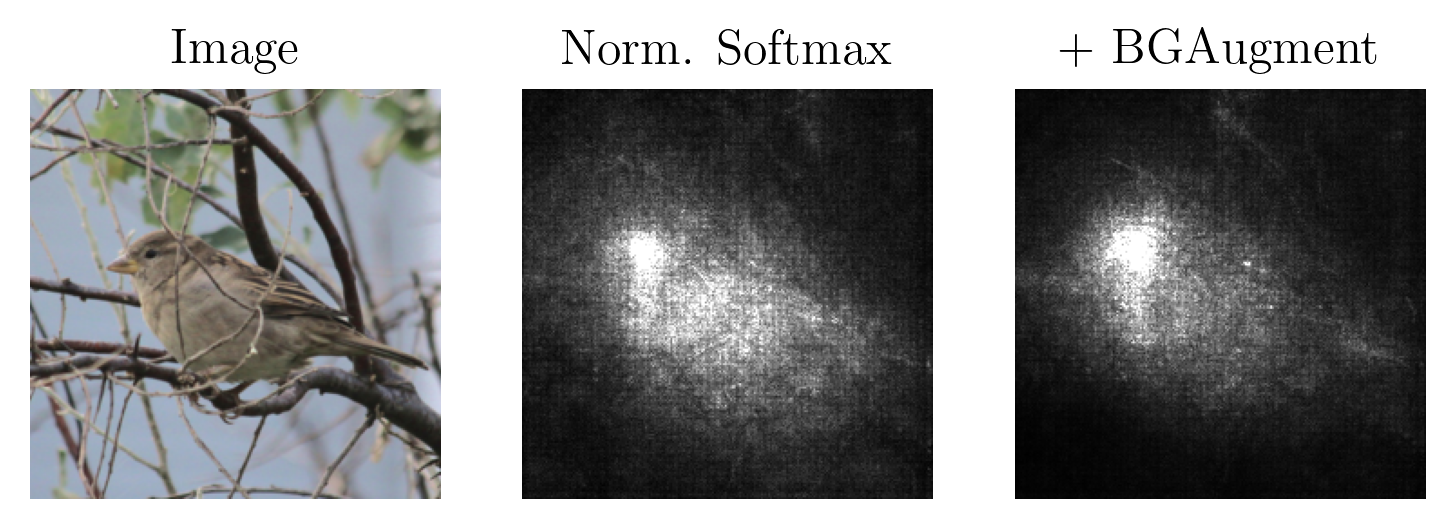}
    \\
    \includegraphics[width=0.48\textwidth]{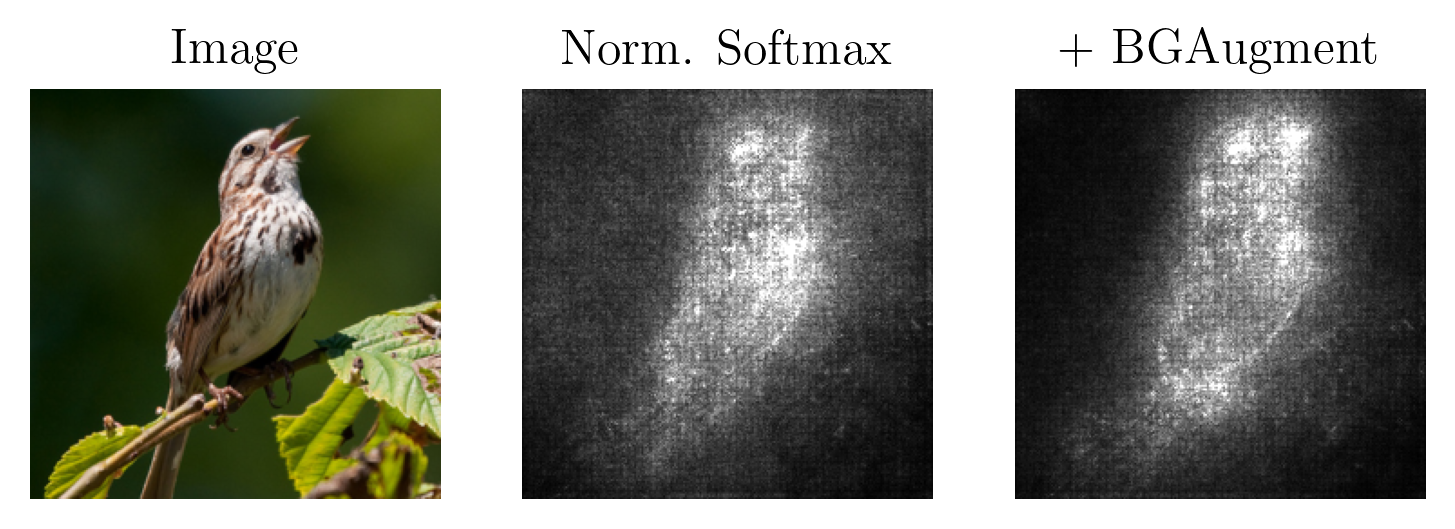}
    \hfill
    \includegraphics[width=0.48\textwidth]{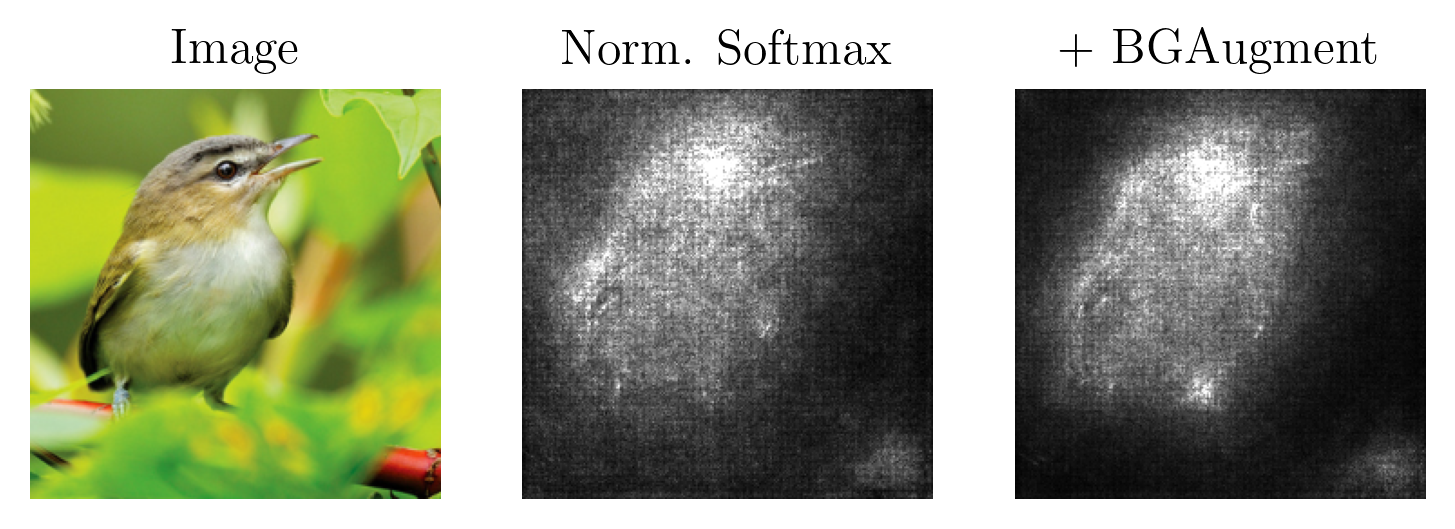}
    \caption{Four images from the CUB200 test set and their corresponding attribution maps (Normalized Softmax Loss model with and without BGAugment). The model focuses more on brighter areas. The base model shows some attention on the background, while the BGAugment model has sharper focus on the main object.}
    \label{fig:qualitative_analysis}
\end{figure}

To better understand the better test performance of BGAugment, we visualize the input pixels the models are most sensitive to using the DML attribution map generation method by Kobs et al.~\cite{kobs2021different}.
\Cref{fig:qualitative_analysis} shows attribution maps of the Normalized Softmax Loss model trained with and without BGAugment.
BGAugment has a darker background area and thus focuses more on the main object in the image.

To quantify this, we introduce a new metric that measures how much of the model's focus is on the main object.
Given a ground truth mask $M \in [0,1]^{h \times w}$ and an attribution map $A \in \mathbb{R}_+^{h \times w}$ of a model for an image $I \in \mathbb{R}^{h \times w \times c}$.
Our ideal metric has the following desired properties:
(1) The best possible value is one, i.e. the model attends to only the foreground area.
(2) If the model does not focus on anything but distributes its attribution uniformly, the metric's value should be zero.
This makes it possible to compare two images where the foreground areas are of different size.
If, for example, the main object has double the pixel count, then a naive metric that measures the percentage of attribution that is on the foreground is also doubled when the attribution is in fact uniformly distributed.
We thus need to account for the size of the foreground object in the metric.
Overall, we propose the following score:
\begin{align}
    f(M) &= \sum_{i,j} \frac{M_{i,j}}{w \cdot h} & \text{// percentage of \underline{f}oreground} \\
    a(M, A) &= \frac{\sum_{i,j} M_{i,j} \cdot A_{i,j}}{\sum_{i,j} A_{i,j}} & \text{// percentage of \underline{a}ttribution on the foreground} \\
    \text{score}(M, A) &= \frac{a(M, A) - f(M)}{1.0 - f(M)} & \text{// final score normalized by foreground size} \label{eq:score}
\end{align}

Our previous experiments have shown the dependence of models on the background only indirectly, by showing that their performance drops substantially when replacing backgrounds.
With this metric, we can directly quantify the attribution that the model assigns to the foreground.
Lower values thus signal more dependence on the background.
If the metric is negative, the model attends more to the background than the foreground.
Overall, the metric can achieve values in the interval $(-\infty, 1]$.
We apply it to all trained models and test datasets and show means and standard deviations in \Cref{tab:analysis}.
It shows that models trained with BGAugment achieve higher values than their basic training counterparts, i.e. are less dependent on the background.
Overall, however, none of the mean scores is negative, meaning that models trained without BGAugment are also focusing on the foreground for the most part.

\begin{table}[t]
    \centering
    \caption{Means and standard deviations for our analysis using our proposed metric (\Cref{eq:score}) to quantify the attribution of the model on the foreground. Lower values indicate larger dependence on the background.}
    \begin{tabular}{lccc}
        \toprule
        {}                          & \textbf{Cars196} & \textbf{CUB200} & \textbf{SOP} \\
        \midrule
        \textbf{Contrastive}        & 0.50 ± 0.09 &  0.25 ± 0.08 &  0.07 ± 0.16 \\
        \textbf{+ BGAugment}        & \textbf{0.53 ± 0.10} &  \textbf{0.32 ± 0.10} &  \textbf{0.31 ± 0.20} \\
        \midrule
        \textbf{Triplet}            & 0.50 ± 0.09 &  0.24 ± 0.08 &  0.08 ± 0.17 \\
        \textbf{+ BGAugment}        & \textbf{0.54 ± 0.09} &  \textbf{0.33 ± 0.09} &  \textbf{0.26 ± 0.22} \\
        \midrule
        \textbf{Multi Similarity}   & \textbf{0.52 ± 0.09} &  0.26 ± 0.08 &  0.11 ± 0.17 \\
        \textbf{+ BGAugment}        & 0.50 ± 0.09 &  \textbf{0.30 ± 0.09} &  \textbf{0.28 ± 0.19} \\
        \midrule
        \textbf{ArcFace}            & 0.49 ± 0.10 &  0.25 ± 0.08 &  0.13 ± 0.14 \\
        \textbf{+ BGAugment}        & \textbf{0.56 ± 0.10} &  \textbf{0.33 ± 0.09} &  \textbf{0.26 ± 0.17} \\
        \midrule
        \textbf{Normalized Softmax} & 0.51 ± 0.09 &  0.24 ± 0.08 &  0.12 ± 0.14 \\
        \textbf{+ BGAugment}        & \textbf{0.56 ± 0.09} &  \textbf{0.30 ± 0.08} &  \textbf{0.28 ± 0.20} \\
        \bottomrule
    \end{tabular}
    \label{tab:analysis}
\end{table}

\section{DISCUSSION \& CONCLUSION}

In this paper, we have shown that Deep Metric Learning suffers from background bias.
Our experiments show that performance can drop substantially when backgrounds are exchanged.
Exchanging image backgrounds during training using a salient object detection network improves performance while having neither model changes, additional parameters, nor increased inference time.
Our qualitative and quantitative analyses confirm that models trained this way focus more on the foreground.
While automatically generated masks from the state-of-the-art salient object detection network mostly isolate the relevant object, masking accuracy is not perfect.
Our hand-annotated samples show that up to \SI{10}{\%} of the ground truth foreground is not present in the generated mask.
While this might partially explain the performance drop in our experiments, we can observe the dependence of standard trained models on the background in \Cref{fig:qualitative_analysis} without needing to trust the mask generation process.

Investigating and combating background bias in DML is beneficial to the development of retrieval settings such as item retrieval or person reidentification systems.
Our work suggests that such systems need to be trained carefully in order to find relevant images without simply relying on unimportant background information.
While background augmentation during training is a viable option to mitigate background bias, adapting other strategies from classification networks to DML is an interesting research direction.
Also, investigating dataset properties that influence background bias is certainly helpful.
With this knowledge, guidelines for a more careful dataset collection can be formulated.

\bibliography{report} %
\bibliographystyle{spiebib} %

\end{document}